\documentclass[11pt]{article}

\usepackage[final]{acl}

\usepackage{times}
\usepackage{latexsym}

\usepackage[T1]{fontenc}
\usepackage[utf8]{inputenc}

\usepackage{microtype}

\usepackage{inconsolata}

\usepackage{graphicx}

\usepackage{booktabs}

\title{Exploring Profiles of Cognitive Distortions Associated with Mental Health Disorders}

\author{Alina Anikejeva \\
  Institute of Computer Science \\
  University of Tartu \\
  Tartu, Estonia \\
  \texttt{alinaanikejeva@gmail.com} \\\And
  Kairit Sirts \\
  Institute of Computer Science \\
  University of Tartu \\
  Tartu, Estonia \\
  \texttt{kairit.sirts@ut.ee} \\}

\begin{document}
\maketitle
\begin{abstract}

Cognitive distortions, distorted patterns of thinking, have been increasingly studied in computational mental health research. Although they are related to many, if not all, mental health disorders, most existing studies focus primarily on depression. In this work, we explore distortion profiles across multiple mental health conditions. We analyzed a large Reddit-based dataset containing posts from nine self-reported mental health groups as well as a control group using both an n-gram-based method and a fine-tuned transformer model for detecting cognitive distortions. Mental health groups, both when pooled together and when examined individually, showed higher prevalence of cognitive distortions compared to the control group, with the effect sizes ranging from small to moderate. When comparing distortion profiles across conditions, we observed largely similar patterns, although some groups exhibited overall higher levels of distortions than others. These findings suggest that relatively simple lexical approaches can be useful for exploratory analyses of group-level trends in large-scale mental health text data.

\end{abstract}

\section{Introduction}

Cognitive distortions (CD) are negative thinking patterns that lead individuals to perceive reality in a
distorted way, affecting thoughts, emotions, and behaviors \citep{beck1979cognitive}. They are often associated with mental health disorders, such as
depression and anxiety \citep{joormann2016examining,kuru2018cognitive,ouhmad2024cognitive}, and may contribute to their development and persistence \citep{burns1999feeling}. 

Previous computational work has primarily focused on detecting cognitive distortions in text using supervised learning methods such as transformer models trained on annotated data \citep{simms2017detecting, shickel2020automatic, shreevastava2021detecting, tauscher2023automated}, as well as more recent approaches based on prompting large language models \citep{chen2023empowering,lim2024erd}. These approaches aim to support clinically relevant predictions at the level of individual users \citep{wang2023cognitive,tauscher2025automated}.

A complementary line of work examines cognitive distortions at the population level by analyzing their prevalence in large-scale text data. In this line of work, \citet{bathina2021depression} introduced a lexicon of distortion-related n-grams and used it to compare distortion prevalence in depression-related texts. The same n-gram-based approach has been applied to anxiety \citep{rutter2025anxiety}, showing associations between distortion frequency and symptom severity. Such studies enable analysis at a scale that is difficult to achieve in traditional clinical settings.

However, these studies have an important methodological limitation. The n-gram-based approach relies on surface-level lexical matches and may produce false positives, as many markers (e.g., ``I am always'') also occur in non-distorted contexts. This raises questions about the validity of the detected patterns and motivates exploratory comparison with more contextual modeling approaches.
In addition, prior work has focused primarily on overall distortion prevalence, without examining how specific distortion categories vary across mental health conditions. 

To our knowledge, the only prior study examining profiles of cognitive distortions is \citet{agarwal2025exploratory}. However, that work focuses on relationships between cognitive distortions and emotion appraisal dimensions, which is a different psychological phenomenon and not directly related to mental health disorders.

In this work, we build on this line of work in two directions. Cognitive distortions are not exclusive to mental health conditions, but represent common patterns of human thinking \citep{beck1979cognitive}. For this reason, we include a control group to provide a baseline for interpreting differences in distortion patterns. We apply the n-gram-based detection approach to a large-scale Reddit dataset covering multiple mental health groups and a control group \citep{cohan2018smhd}. We analyze distortion patterns across groups to determine whether CD profiles differ across disorders or remain largely similar. 
To provide secondary comparison to the n-gram-based approach, we additionally train a transformer-based model and compare whether the observed patterns are consistent across both methods.
In summary, this study addresses the following research questions:

Q1: Are there differences in cognitive distortion prevalence between mental health groups and the control group?

Q2: How do distortion patterns vary across different mental health groups?

Q3: Are the patterns produced by the n-gram and transformer-based approaches consistent?

\section{Data}
We use three datasets in our work: 1) a large Reddit-based dataset to study CD profiles, 2) a small dataset annotated with CD labels to train supervised models to predict CD labels, and 3) a lexicon of CD-related n-grams. This section describes all these three datasets.

\subsection{SMHD dataset}

We use the Self-reported Mental Health Diagnoses (SMHD) dataset \citep{cohan2018smhd} for the analysis of CDs. SMHD contains Reddit posts and comments from users who explicitly self-reported having been diagnosed with one or more mental health conditions from nine diagnostic categories. The dataset includes user-level labels indicating reported diagnoses and contains posts made between January 2006 and December 2017.

The diagnosed users (Clinical Group) were identified using pattern-based matching of explicit diagnosis statements (e.g., ``I was officially diagnosed with OCD last year''). Users could be assigned multiple conditions if they reported more than one diagnosis. The control group was selected from users without mental health posts and matched to diagnosed users based on posting activity. Specifically, control users were required to post in the same subreddits and have a comparable number of posts to ensure similar activity patterns.

To reduce topic-related bias, the SMHD dataset excludes posts related to mental health for diagnosed users. Mental health posts were defined as posts made in mental health–related subreddits or containing mental health–specific language. As a result, the dataset consists of general, non-mental-health-specific content.

\subsubsection{Preprocessing}

We removed users with multiple mental health labels  to ensure mutually exclusive group assignments. We retained only Reddit posts (excluding comments), as posts generally provide longer and more contextually complete text.

Text was cleaned using standard preprocessing steps, including lowercase conversion, whitespace normalization, removal of formatting characters, and filtering to English-language posts. We also applied Reddit-specific noise filtering by removing trading or advertising content and URL-heavy posts (>25\% URLs).

After preprocessing, the control group was substantially larger than any individual diagnostic group. To address this imbalance, we randomly generated 30 control subsets, each matched in size to the largest diagnostic group (ADHD; 3,649 users). As shown in Table~\ref{tab:dataset_stats}, 109,470 control users were randomly sampled and evenly divided into 30 distinct control subsets. 

\begin{table}[t]
\centering
\small
\begin{tabular}{lrr}
\toprule
\textbf{Label} & \textbf{Total Users} & \textbf{Total Posts} \\
\midrule
Control (30) & 109,470 & 1,981,786 \\
\midrule
ADHD & 3,649 & 47,052 \\
Depression & 3,647 & 45,362 \\
Bipolar & 1,932 & 21,834 \\
Anxiety & 1,633 & 20,708 \\
Autism & 955 & 13,480 \\
PTSD & 795 & 8,670 \\
OCD & 581 & 7,403 \\
Schizophrenia & 441 & 5,920 \\
Eating Disorder & 183 & 1,649 \\
\bottomrule
\end{tabular}
\caption{Number of users and posts per group after preprocessing. The control group was divided into 30 equally sized, non-overlapping subsets matched to the ADHD group.}
\label{tab:dataset_stats}
\end{table}

\subsection{Therapist Q\&A dataset}

The supervised transformer model to predict CD labels in the SMHD dataset was trained on the Therapist Q\&A (QA) dataset. The dataset was sourced from Kaggle and annotated as part of a prior study for cognitive distortion detection with ten CD categories \citep{shreevastava2021detecting}. The dataset consists of 2530 anonymized therapist–patient question–answer pairs, which was divided into training, development and test sets using 70/10/20 splits. Only the patient input was used in this work to train the models.

\subsection{N-grams dataset}

For the lexicon-based detection approach, we use the n-gram list introduced in prior work on cognitive distortion identification \citep{bathina2021depression}, compiled by a panel of mental health professionals and publicly available in the corresponding GitHub repository.\footnote{\url{https://github.com/mctenthij/CDS_paper}} The resource includes twelve cognitive distortion categories and associated linguistic markers (n-grams), along with their variants.

All n-grams were lowercased to align with text preprocessing. The number of markers varies across categories (e.g., Mindreading includes substantially more markers than Emotional Reasoning), and some categories rely on more common lexical items (e.g., “all,” “nothing”) than others. This may influence raw match frequencies; comparisons are therefore interpreted relative to group-level distributions. The n-gram categories and their marker counts are presented in Table~\ref{tab:ngram_distribution}. 

\begin{table}[ht]
\centering
\small
\setlength{\tabcolsep}{2.4pt}
\begin{tabular}{lr}
\toprule
\textbf{Category} & \textbf{Total N-grams} \\
\midrule
Mindreading & 159 \\
Labeling and mislabeling & 68 \\
Overgeneralizing & 31 \\
Dichotomous Reasoning & 24 \\
Fortune-telling & 24 \\
Catastrophizing & 23 \\
Disqualifying the Positive & 19 \\
Personalizing & 16 \\
Mental Filtering & 14 \\
Magnification and Minimization & 11 \\
Should statements & 11 \\
Emotional Reasoning & 7 \\
\bottomrule
\end{tabular}
\caption{Distribution of n-grams across cognitive distortion categories.}
\label{tab:ngram_distribution}
\end{table}

\section{Methods}

In this section, we describe how cognitive distortion profiles are computed for mental health groups. We define a profile as the vector of effect sizes across distortion categories for a given group. We begin with an overview of the analysis pipeline, followed by the n-gram–based detection method, the procedure for establishing control-based baselines, and the transformer-based model used as a secondary comparison based on more contextual representations than the n-grams.

\subsection{Overall process}
\label{subsec:overall}
After preprocessing, cognitive distortions were identified and analyzed across diagnostic groups in four stages:
\begin{enumerate}
    \item Each post was matched against a predefined set of n-gram markers representing twelve CD categories, and category-specific marker counts were computed.
    \item Distortion frequency distributions were estimated using the control subgroups. Percentile-based thresholds derived from these distributions were then used to determine whether a post exhibited elevated levels of a given distortion.
    \item Statistical tests were conducted to examine differences in distortion prevalence between self-diagnosed individuals and clinical controls, as well as across self-diagnosed diagnostic groups.
    \item Two pretrained transformer models (BERT and RoBERTa) were evaluated on an annotated QA dataset using macro F1. The best-performing model was subsequently applied to the SMHD corpus to explore whether broadly similar aggregate-level tendencies emerge under a substantially different detection approach.
\end{enumerate} 

\begin{figure}[htbp]
    \centering
    \includegraphics[width=0.8\linewidth]{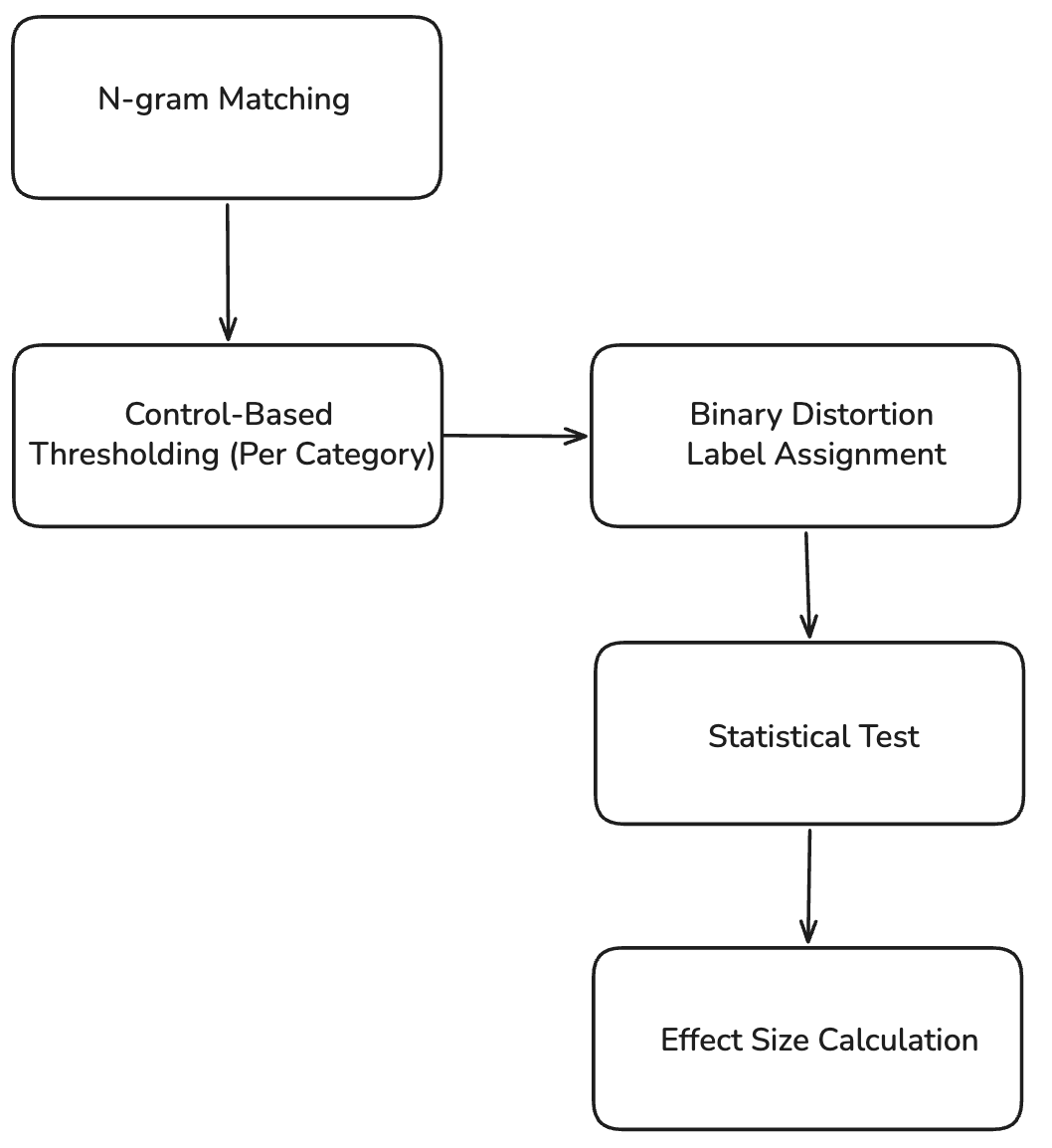}
    \caption{Overview of the analysis steps from n-gram matching through thresholding and binary labeling to statistical testing.}
    \label{fig:pipeline}
\end{figure}

Figure~\ref{fig:pipeline}. provides a schematic overview of the analysis steps, from n-gram matching to thresholding, binary labeling, and statistical testing.

\subsection{N-gram Matching}

Cognitive distortions were identified using a predefined lexicon of n-gram markers and their variants, each assigned to a specific distortion category. Posts were matched using exact, case-insensitive string matching after lowercasing both markers and text. For each post, occurrences were counted per category.

\subsection{Thresholding}

Of the 2,153,864 posts, 57.7\% contained at least one n-gram match, reflecting the fact that many markers also occur in everyday language. To reduce false positives, category-specific thresholds were derived from control-group distributions. 
For each distortion category, the 75th percentile of control marker counts was used as a cutoff, filtering typical language while retaining posts with elevated distortion-related frequency. The 75th percentile was chosen as a conservative cutoff to filter common lexical occurrences while retaining posts with relatively elevated distortion-related frequency. 
Alternative percentile cutoffs (50th, 90th, and 95th percentiles) were also explored to examine how threshold values vary across distortion categories (Appendix~\ref{app:thresholding}).
For the majority of categories, the 75th percentile exceeded one match; however, Dichotomous Reasoning, Labeling and Mislabeling, and Should Statements required thresholds of 3, 2, and 2 matches, respectively. 

Posts exceeding the category-specific threshold were labeled as distorted for that category as binary label; posts exceeding at least one category threshold were labeled as distorted overall.

\subsection{Statistical Testing}
Based on the binary labels, proportions of distorted posts were calculated to enable fair comparisons across groups with unequal post counts. Group differences were evaluated using two-tailed two-sample z-tests for proportions at $\alpha$ = 0.01. Tests were conducted both at the aggregate level (all self-reported diagnoses vs. controls) and at the subgroup level, where each diagnostic group was compared against its 30 matched control subgroups. To control for multiple comparisons, Holm–Bonferroni correction was applied.

To quantify the magnitude of differences, effect sizes were calculated using Cohen’s h, which measures standardized differences between proportions. Median Cohen’s h values were then aggregated across comparisons to construct distortion profiles for each subgroup.

\subsection{Transformer model}

BERT and RoBERTa models pretrained on mental health data \citep{ji2022mentalbert} were fine-tuned on the annotated multi-label QA dataset \citep{shreevastava2021detecting}.
Because each input may contain multiple cognitive distortions, the task was formulated as a multi-label classification problem, with sigmoid activation used to produce independent probability scores for each label. Models were trained using binary cross-entropy loss and evaluated using macro F1 across five runs to account for variability.

During inference, probability scores were converted into binary predictions using a decision threshold optimized on the validation set to maximize macro F1. Based on this evaluation, the better-performing model was selected and subsequently applied to the SMHD corpus. Because the QA dataset uses a different cognitive distortion taxonomy than the n-gram lexicon, model predictions were generated according to the QA label set.

\section{Results}
We first present results obtained using the n-gram–based method, followed by results based on labels produced by the transformer models.

\subsection{N-gram matching}
In this section, we present results based on the n-gram–based detection method. We first examine overall differences in distortion prevalence between Clinical and Control groups, and then analyze variation across cognitive distortion categories and distortion profiles across conditions.

\subsubsection{Distorted vs Non-Distorted}

\paragraph{Overall comparison}
Using the 75th percentile threshold, 29.8\% of all posts were labeled as distorted, with a higher proportion observed in the Clinical group (44.5\%) than in the Control group (28.5\%) . This difference was statistically significant (two-tailed z-test, p < 0.01) and corresponded to a small-to-moderate effect size (Cohen’s $h$ = 0.34), indicating that cognitive distortions are more common in the Clinical group compared to the Control group.

\begin{figure}[htbp]
\centering
\includegraphics[width=1\linewidth]{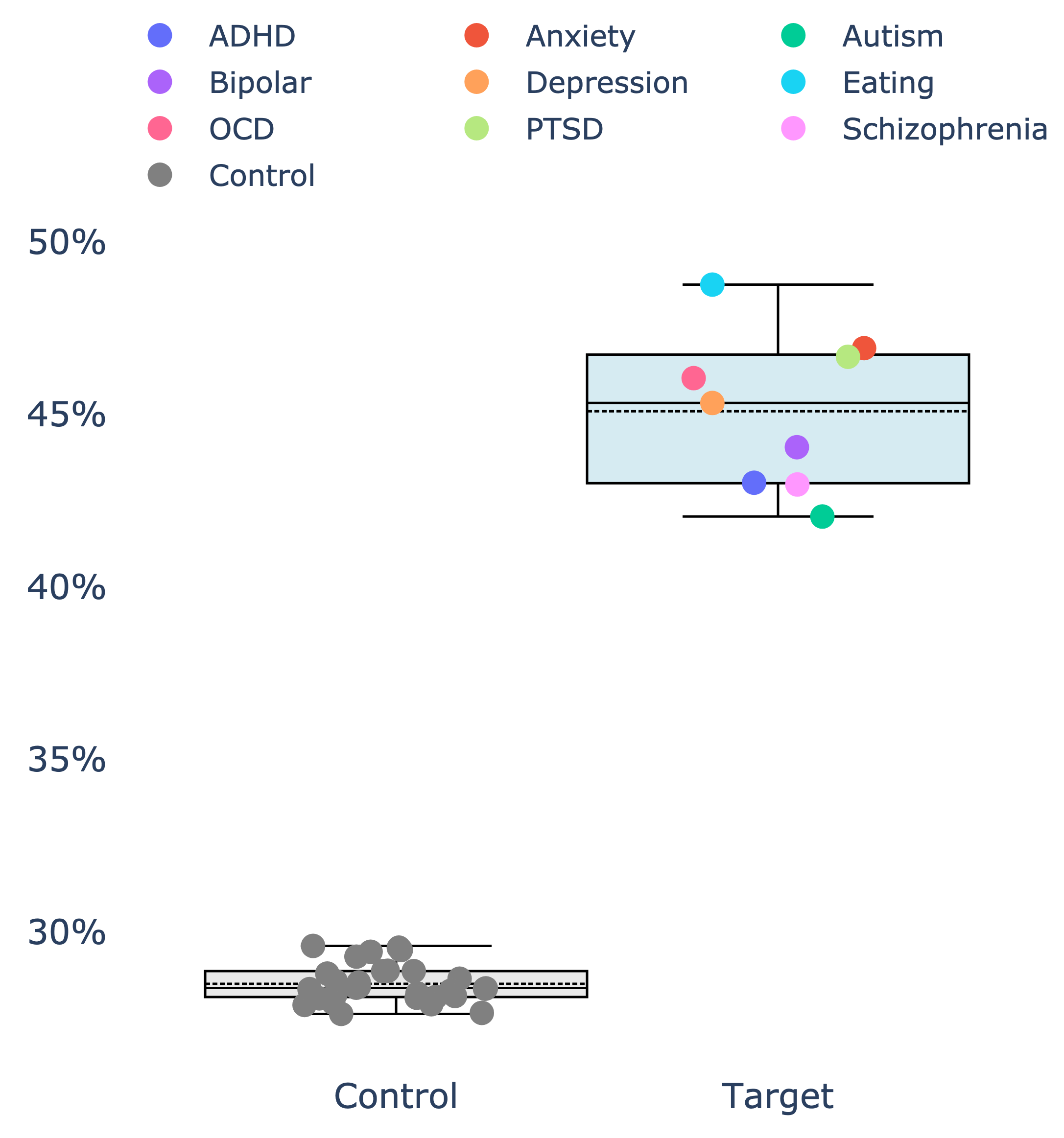}
\caption{Distribution of distorted post percentages across control and clinical groups. Boxplots show group distributions, with individual points representing subgroup-level values.}
\label{fig:distortion_plot}
\end{figure}

\paragraph{Individual Clinical group comparison}
We compared distortion rates between individual Clinical and Control subgroups. Control subgroups consistently showed lower distortion rates (28--30\%) than Clinical subgroups (43--48\%) (Figure~\ref{fig:distortion_plot}). All Clinical groups differed significantly from matched Controls after Holm--Bonferroni correction, with small-to-moderate and relatively consistent median effect sizes ($h = 0.29$--$0.42$), highest for Eating Disorder and lowest for Autism (Table~\ref{tab:cohen_h_summary}).
These results show that the difference between Clinical and Control groups persists across specific disorder categories.

\begin{table}[h]
\centering
\small
\setlength{\tabcolsep}{5.7pt}
\begin{tabular}{lccc}
\toprule
\textbf{Group} & \textbf{Min $h$} & \textbf{Median $h$} & \textbf{Max $h$} \\
\midrule
Schizophrenia & 0.28 & 0.31 & 0.32 \\
PTSD & 0.35 & 0.38 & 0.40 \\
OCD & 0.34 & 0.37 & 0.38 \\
Eating Disorder & 0.40 & 0.42 & 0.44 \\
Depression & 0.33 & 0.35 & 0.37 \\
Bipolar & 0.30 & 0.33 & 0.34 \\
Autism & 0.26 & 0.29 & 0.30 \\
Anxiety & 0.36 & 0.39 & 0.40 \\
ADHD & 0.28 & 0.31 & 0.32 \\
\bottomrule
\end{tabular}
\caption{Minimum, median and maximum Cohen’s $h$ effect sizes by diagnostic group}
\label{tab:cohen_h_summary}
\end{table}

\begin{table}[h]
\centering
\small
\begin{tabular}{lc}
\toprule
\textbf{Distortion Category} & \textbf{Cohen’s $h$} \\
\midrule
Dichotomous Reasoning & 0.34 \\
Labeling and Mislabeling & 0.25 \\
Personalizing & 0.25 \\
Should Statements & 0.19 \\
Overgeneralizing & 0.17 \\
Magnification and Minimization & 0.11 \\
Mindreading & 0.11 \\
Fortune-telling & 0.10 \\
Emotional Reasoning & 0.09 \\
Disqualifying the Positive & 0.08 \\
Mental Filtering & 0.03 \\
Catastrophizing & 0.02 \\
\bottomrule
\end{tabular}
\caption{Cohen’s $h$ Effect Sizes by Cognitive Distortion Category}
\label{tab:cohen_h_categories}
\end{table}

\subsubsection{Variations in Cognitive Distortions}

\paragraph{Overall comparison}
We compared distortion percentages across cognitive distortion categories between the pooled Clinical and Control groups. Overall, the Clinical group exhibited higher distortion rates than the Control group across all categories. Some distortions were relatively common (e.g., Dichotomous Reasoning: ~33\% in Clinical vs.~15\% in Control), whereas others were rare in both groups (e.g., Catastrophizing: <1\%). The largest differences were observed for Dichotomous Reasoning, Labeling, Should Statements, and Personalizing, with smaller gaps for Mindreading and Fortune-telling (see Appendix~\ref{app:category_ngrams}).

Two-tailed two-proportion z-tests confirmed that these differences were statistically significant across all categories, with generally small effect sizes and relatively larger (small-to-moderate) effects for Dichotomous Reasoning, Labeling and Mislabeling, and Personalizing (Table~\ref{tab:cohen_h_categories}).
This indicates that while distortions are more frequent in the Clinical group across all categories, the magnitude of these differences varies by distortion type.

\paragraph{Individual Clinical group comparison}
We examined variation in distortion prevalence and effect size profiles across individual Clinical subgroups compared to Control subgroups. Distortion prevalence varied across Clinical subgroups, with Eating Disorder and PTSD showing higher percentages across multiple categories, while other Clinical groups exhibited lower—but still elevated—rates relative to Controls. Variation between Clinical groups differed by distortion type, with some categories showing consistent patterns and others greater dispersion.

\begin{figure}[t]
    \centering
    \includegraphics[width=\linewidth]{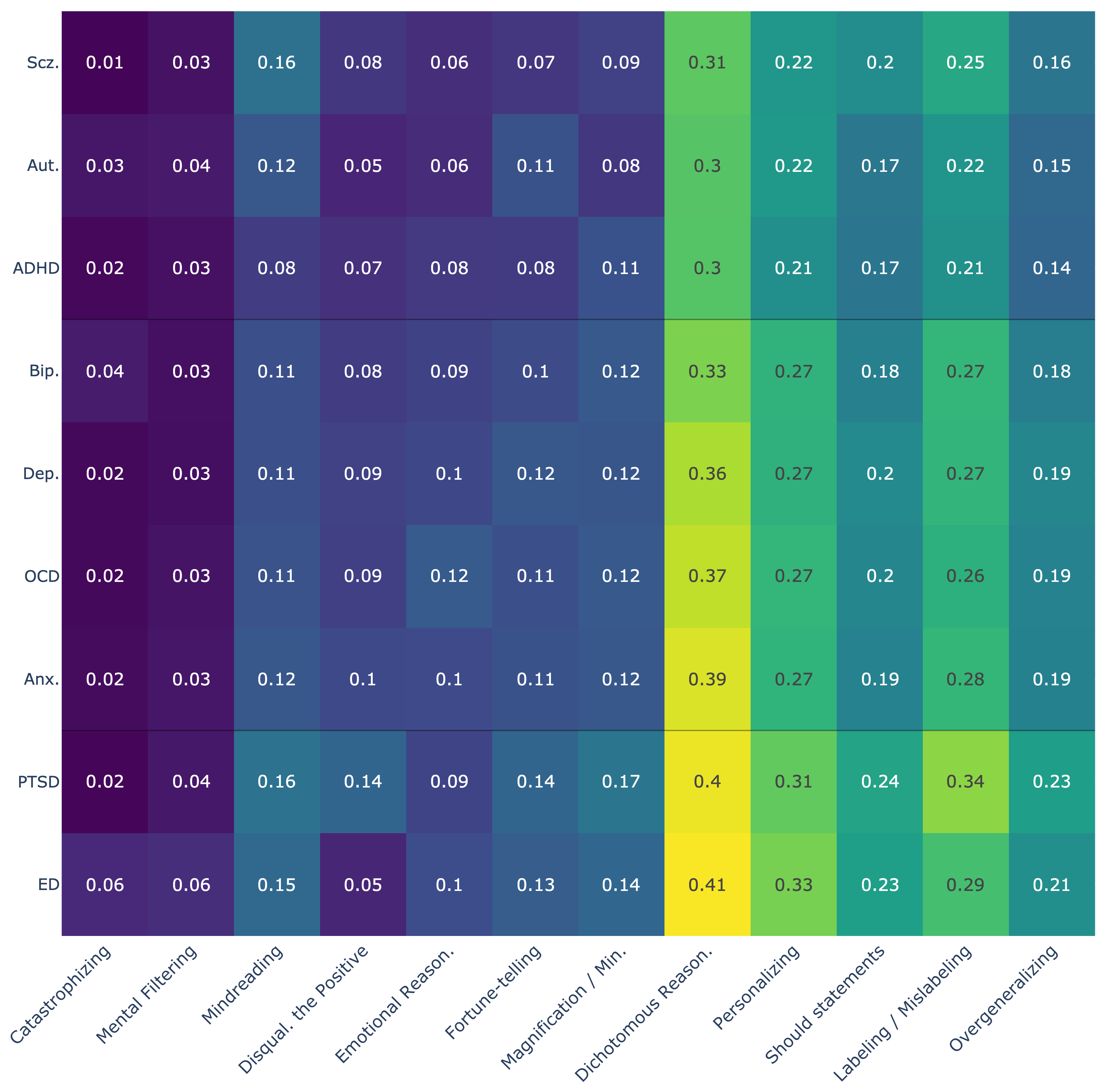}
    \caption{
    Cognitive distortion profiles across mental health conditions. 
    Cells display median Cohen’s $h$ effect sizes for each distortion–condition pair. 
    Abbreviations: ED = Eating Disorder, Anx. = Anxiety, Dep. = Depression, 
    Bip. = Bipolar Disorder, Aut. = Autism, SCZ = Schizophrenia.
    }
    \label{fig:distortion_heatmap}
\end{figure}

Across all combinations of distortion categories, Clinical subgroups, and Control subgroups, 3,240 two-proportion z-tests were conducted; after Holm--Bonferroni correction, 2,835 remained statistically significant and 405 were not. Corresponding effect sizes were computed for each comparison, and median Cohen’s $h$ values were used to construct distortion profiles for each disorder, as shown in Figure~\ref{fig:distortion_heatmap}.

Distortion profiles showed limited separation between disorders, with effect size differences across categories generally not exceeding 0.10--0.15. Weak grouping patterns emerged: Eating Disorder and PTSD exhibited the highest median effects across most categories; Anxiety, Depression, OCD, and Bipolar showed intermediate effects; and ADHD, Autism, and Schizophrenia showed the lowest.

Across all groups, the relative ranking of distortion categories was similar, with Dichotomous Reasoning consistently showing the largest effects, followed by Personalizing and Labeling, while Mental Filtering, Catastrophizing, and Disqualifying the Positive showed the smallest effects.

Overall, these results indicate that while distortion prevalence differs across Clinical subgroups, the overall pattern of cognitive distortions is largely shared, with only modest differentiation between disorders.

\subsection{Transformer model}

In this section, we present results based on the transformer-based detection method. Unlike the n-gram approach, which serves as the primary method, the transformer model is used to assess whether the group-level patterns observed in the lexical analysis are reproduced by a more contextual detection approach. We first report model performance, followed by the analysis of distortion prevalence across groups.

\subsubsection{Model Accuracy}
The transformer model achieved macro F1 scores of approximately 0.25--0.29 on the development set, but substantially lower performance on the held-out QA test set ($\approx$ 0.07--0.08), suggesting instability in fine-grained multi-label category prediction. One possible contributing factor is overfitting related to the decision threshold selection procedure adopted in the multi-label setting.

At the same time, binary distorted vs non-distorted classification performance (derived from the same multi-label predictions) was considerably higher on the test set (F1 = 0.56 for BERT and 0.75 for RoBERTa), indicating that the models capture broader distortion-related signal more reliably than individual distortion categories.

Previous work reporting F1 scores in the range of 0.2--0.3 \citep{chen2023empowering,lim2024erd} formulates the task as multi-class classification, whereas our setup uses multi-label formulation, making the results not directly comparable. In addition, the transformer models were trained on patient questions in the therapist-patient interactions discussing mental health topics, but later applied to general Reddit posts from the SMHD dataset, introducing substantial domain differences between training and application data. Because of these limitations, the transformer-based analysis should not be interpreted as reliable fine-grained cognitive distortion detection at the individual level. Instead, it serves as a secondary exploratory comparison examining whether broadly similar tendencies emerge under a substantially different modeling approach.

\subsubsection{Distorted vs Non-Distorted}

\paragraph{Overall comparison}
Using the RoBERTa model, 3.1\% of posts were labeled as distorted. The proportion was 9.4\% in the Clinical group and 2.5\% in the Control group. 

\paragraph{Individual Clinical group comparison}
We examined variation in transformer-identified distortion rates across individual Clinical subgroups compared to Control subgroups. Control subgroups showed stable distortion rates of approximately 3\%, whereas all Clinical subgroups exhibited higher proportions (Figure~\ref{fig:transformer_results}). Eating Disorder and PTSD displayed the highest distortion rates among Clinical groups.

\subsubsection{Variations in Cognitive Distortions}
Across cognitive distortion categories, posts in Clinical subgroups were more frequently labeled as distorted than those in the Control group, with Should Statements showing the highest prevalence overall. Eating Disorder consistently appeared near the upper range across categories, whereas ADHD tended to appear near the lower range (see Appendix~\ref{app:category_trans}).
This indicates that the model is capturing consistent differences between Clinical and Control groups, with variation across disorders but similar category-level patterns.

\subsection{Comparison of N-Gram Method and Transformer Model}
\begin{figure}[htbp]
\centering
\includegraphics[width=1\linewidth]{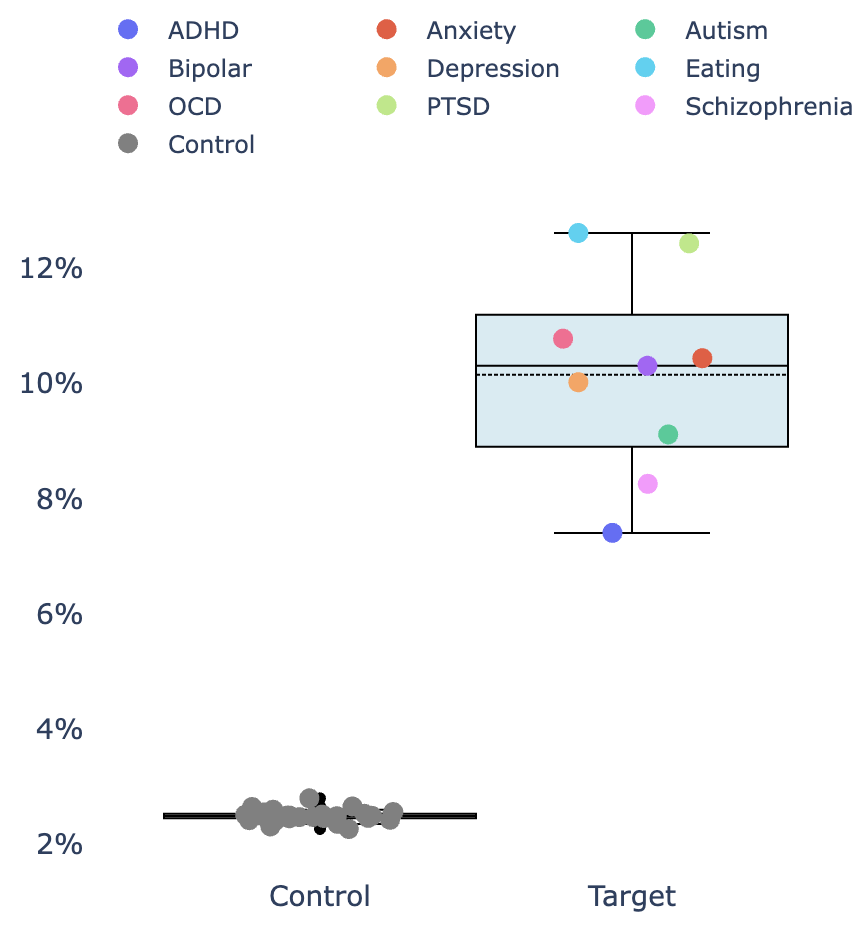}
\caption{Transformer-based distortion rates across control and clinical subgroups.}
\label{fig:transformer_results}
\end{figure}

Although the transformer models identified a considerably lower proportion of distorted posts in both Clinical and Control groups, the overall patterns are similar across both methods. When comparing Figures~\ref{fig:distortion_plot} and~\ref{fig:transformer_results}, one can see a similar pattern, with Eating disorder having the highest proportion of cognitive distortions with both methods, followed by PTSD (although with the n-gram method, Anxiety is on the same level with PTSD), while with both methods, ADHD, Schizophrenia and Autism showing the least proportion of cognitive distortions. 

Additional visualizations for each distortion category and disorder are provided in the Appendix. While these plots are not analyzed in detail here, they show broadly similar patterns across methods, suggesting partial convergence between the two approaches at the group level.

\section{Discussion}

This work set out to examine the profiles of cognitive distortions across different mental health disorders. The analysis was conducted using two methods for distortion detection, with an n-gram approach as the primary technique and a transformer model used as a secondary exploratory comparison based on contextual representations.
The study specifically investigated three questions: whether there are differences in cognitive distortion prevalence between mental health groups and the control group, how distortion profiles vary across conditions, and the extent to which the two methodological approaches produce convergent findings.

First, we found that indeed, cognitive distortions were more prevalent in mental-health-related groups, both when all clinical groups were pooled together as well as when analyzing each group separately. This result aligns with findings from previous studies on depression \citep{bathina2021depression,lalk2025depression} and anxiety \citep{rutter2025anxiety}. The observed effect sizes were generally in the small-to-moderate range, which is perhaps expected because cognitive distortions are essentially normative phenomena and are not exclusive to clinical populations. 
At the same time, our analysis was conducted on a very large social media dataset, where even subtle differences can become statistically significant. Therefore the interpretation of the results should rely more on effect sizes and overall patterns than on statistical significance alone. Although many comparisons remained significant after correction, the observed effect sizes remained relatively modest across most disorder and distortion-category combination.

 When looking at the distinct cognitive distortions, we saw some distortions, particularly Dichotomous Thinking, Labeling and Mislabeling, and Personalizing, that showed larger differences between the mental health and control groups, while some other distortion categories, like Catastrophizing and Mental Filtering showed very small differences between Clinical and Control groups across mental health disorders. These results partially converge with the previous results reported by \citet{bathina2021depression} on depression, where they found that the largest difference between people with depression and a random social media user group was Personalizing, while there were no significant differences between groups related to Catastrophizing and Mental Filtering. Emotional reasoning, which showed the second largest difference in the \citet{bathina2021depression} study, showed only small difference in our research, while in their study, Dichotomous thinking, while significant, was associated with a relatively small difference between groups.

The profiles of cognitive distortions between different mental health disorders were analyzed qualitatively by examining the effect size patterns. Based on this analysis, we observed that the overall structure of the profiles was largely similar across disorders, with differences in effect sizes typically not exceeding 0.10--0.15. Although some variation was present, particularly with Eating Disorder and PTSD groups showing somewhat higher levels of distortions compared to control groups and ADHD and Autism groups appearing closer to the control group, the relative ordering of distortion categories was largely similar across conditions. 
One possible interpretation of these findings is that the observed differences between mental health conditions may be related more to the overall intensity of the distorted language rather than to distinct combinations of the distortion types. 
The elevated distortion levels observed in PTSD and Eating Disorder groups may also partly reflect attempts to establish predictability, control, or emotional regulation in the context of heightened distress, although this interpretation would require substantially more clinically grounded investigation. More generally, these interpretations remain tentative, and more precise methods would be needed to determine whether more distinct disorder-specific cognitive distortion profiles exist.

When comparing the two cognitive distortion detection methods used in this paper, we observed that the transformer-based model detected considerably fewer distortions. Despite these differences, the overall patterns observed were similar across the two methods. Thus, we conclude that although the n-gram-based method is not adequate for making predictions with high precision due to the high rate of false positives, the results suggest it may still be useful for exploratory analysis of broader group-level trends in large datasets, provided that a control group is used to establish a baseline for interpreting the results. In such large-scale settings, applying transformer models is computationally substantially more costly, while the simple n-gram method offers a lightweight alternative.

Overall our findings show that large-scale computational analysis can detect subtle group-level differences, even when the analytical setting (data and methods) is noisy. While a common line of research focuses on predicting mental health states of specific individuals based on their expression of cognitive distortions \citep{wang2023cognitive} or assessing relations between cognitive distortions and disorder symptoms \citep{lalk2025depression,varadarajan2025linking}, our study, similarly to \citep{bathina2021depression}, aims to reveal the broader associations between cognitive distortions and mental health disorders. We assume that the noise stemming from data and methods affects both the Control and Clinical groups in a similar way. Therefore, while our detection methods are too weak to reliably identify cognitive distortions in the texts of individual users or make inferences about their mental health states, they appear sufficient for detecting broader trends. Such trends can help generate more precise hypotheses for future work, which could then be tested using more targeted methods.

\section{Conclusion}
In this work, we analyzed cognitive distortion patterns across multiple mental health conditions using a large-scale Reddit dataset. Posts from mental health groups showed a higher prevalence of distortions compared to the control group, although the observed effect sizes were generally small to moderate. When comparing profiles across disorders, the relative ordering of distortion categories was largely similar, with only modest differences in effect sizes between conditions. 
Although the analysis relied on a noisy n-gram-based detection method, the overall patterns were consistent with those obtained using a fine-tuned transformer model. This suggests that relatively simple detection approaches can be suitable for exploratory analyses aimed at identifying group-level trends in mental health–related language.

\section*{Limitations}
This study has several limitations.
Although the n-gram-based approach used as the main method is very noisy, producing many false positives because the set of n-grams includes many common phrases, we do not consider this to be an inherent limitation of the study per se, as the use of a control group helps establish a normative baseline. However, cognitive distortions are essentially semantic and contextual phenomena, whereas our n-gram-based detection method is lexical. This means that the method cannot distinguish between someone saying ``I always fail at everything'', potentially reflecting a distorted thought, and ``People often say `I always fail' when they are catastrophizing'', which does not reflect a distorted thought. 
The thresholding procedure used to filter common lexical matches can influence absolute prevalence estimates across distortion categories, particularly because categories differ considerably in the number and frequency of associated n-grams. Although alternative percentile thresholds were explored, the study does not systematically evaluate the sensitivity of the downstream results to threshold selection.
Similarly, some categories relied on broad and relatively common lexical items, whereas others were represented by fewer and more specific expressions. This may partly influence relative prevalence estimates and effect size patterns across categories, meaning that comparisons between distortion categories should be interpreted cautiously.
Nevertheless, because the transformer-based model, which is inherently more contextual, revealed broadly similar aggregate-level tendencies, we have reason to believe that this limitation did not substantially affect the main findings.

The second limitation relates to the Reddit-based SMHD dataset used in this study. Although the self-reported mental health diagnoses were extracted using high-precision textual patterns, the labels are still unavoidably noisy. First, the control groups are created based on the assumption that users who did not visit any mental health-related subreddit and did not self-report any mental health issues do not have such issues, which may not necessarily be true. Second, the posts of the users span a timeline of 11 years, and while a person might have had a diagnosis at one point, this might not be reflective---at least not for all disorders analyzed---of the entire time period. However, because the same limitation applies to both the Control and Clinical groups, we believe it does not invalidate the overall results.

In addition, the decision to retain only users with a single diagnostic label simplifies the clinical reality of substantial comorbidity between mental health conditions, particularly for eating disorders. As a result, the Eating Disorder subgroup represents a relatively narrow subset of users and probably does not fully capture broader clinical eating-disorder populations.

Finally, the analysis of cognitive distortion patterns across mental health disorders was essentially qualitative, relying on the examination of effect size patterns. While this approach is suitable for an exploratory comparison, more rigorous statistical or modeling approaches would be needed to establish whether distinct cognitive profiles exist across mental health disorders.

\section*{Acknowledgments}
This research was supported by the Estonian Centre of Excellence in AI (EXAI) and by the Estonian Research Council Grant PRG3182.

\bibliography{custom}
\clearpage

\appendix

\onecolumn

\section{Cutoff values for n-grams}
\label{app:thresholding}
\begin{center}
    \includegraphics[width=\textwidth]{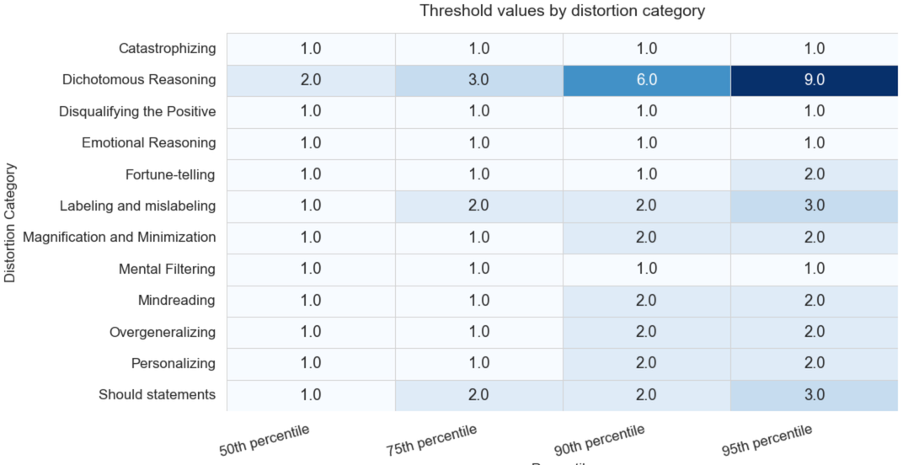}
    \captionof{figure}{Threshold values used for different percentile cutoffs across cognitive distortion categories.}
    \label{fig:category_ngram}
\end{center}

\vspace{35em}

\section{Category-level n-gram results}
\label{app:category_ngrams}

\begin{center}
    \includegraphics[width=\textwidth]{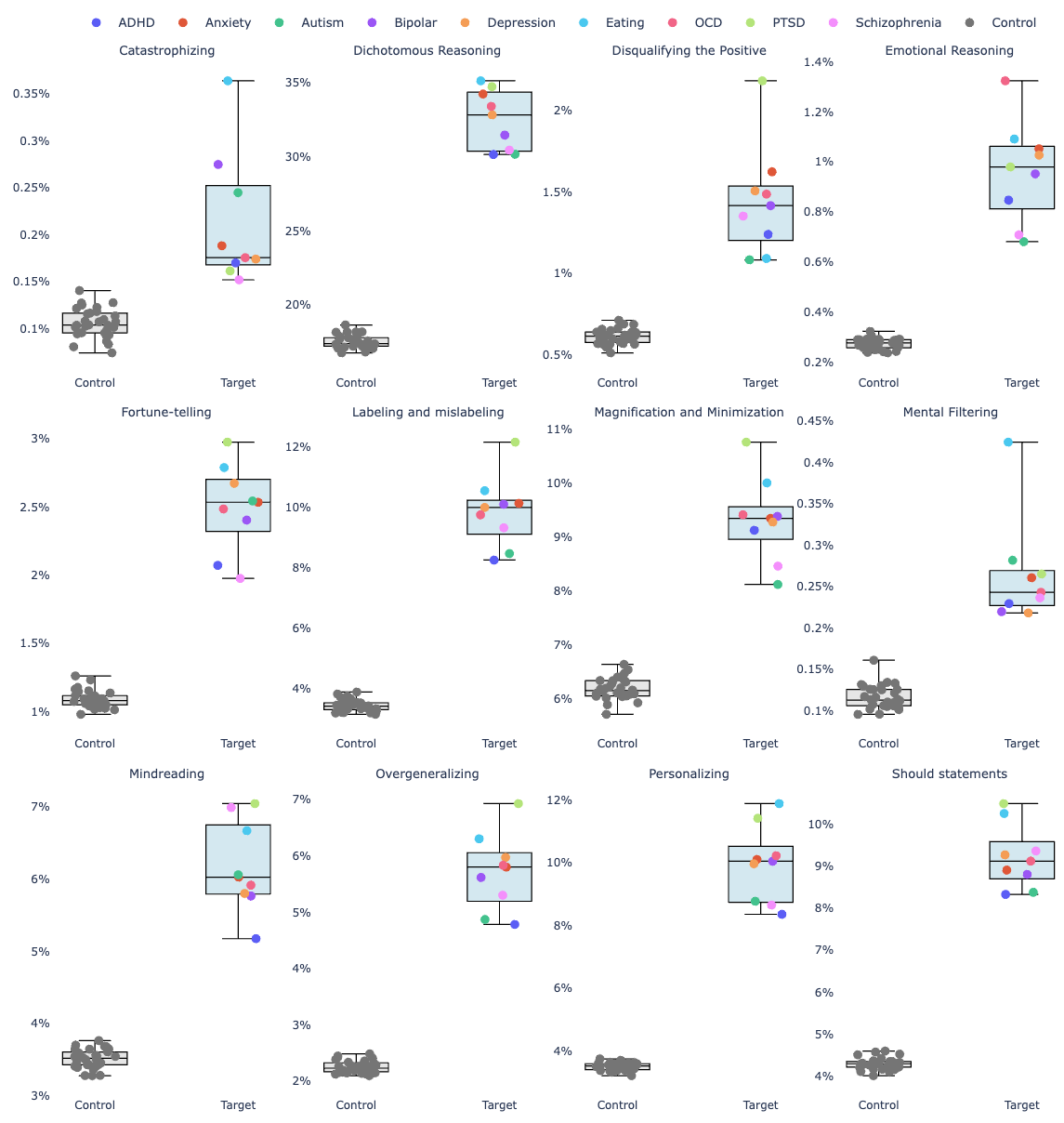}
    \captionof{figure}{Distribution of distortion percentages across cognitive distortion categories for Clinical and Control groups.}
    \label{fig:category_ngram}
\end{center}

\vspace{15em}

\section{Category-level transformer results}
\label{app:category_trans}

\begin{center}
    \includegraphics[width=\textwidth]{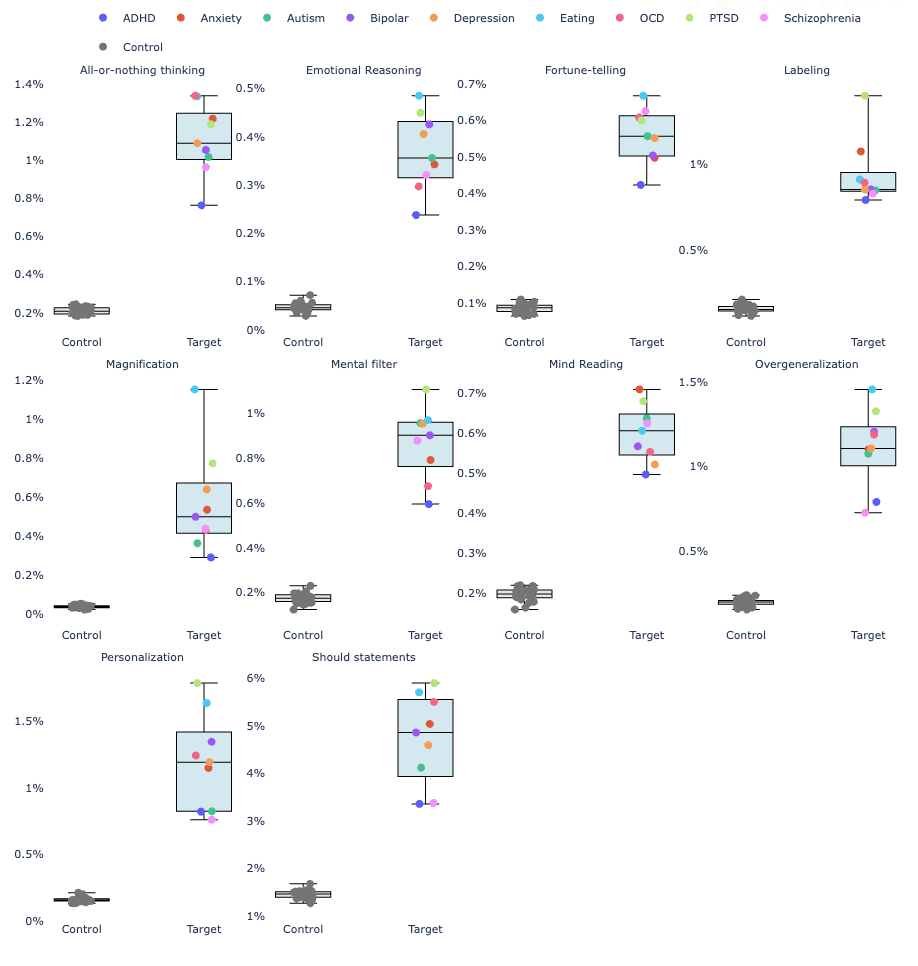}
    \captionof{figure}{Distribution of distortion predictions across cognitive distortion categories based on the transformer model.}
    \label{fig:category_trans}
\end{center}

\end{document}